\documentclass[a4paper,fleqn]{cas-dc}

\usepackage{amsmath,amsfonts,amssymb}
\usepackage{algorithmic}
\usepackage[ruled,linesnumbered]{algorithm2e}
\usepackage{array}
\usepackage[caption=false,font=normalsize,labelfont=sf,textfont=sf]{subfig}
\usepackage{textcomp}
\usepackage{stfloats}
\usepackage{url}
\usepackage{verbatim}
\usepackage{graphicx}
\usepackage{hyperref}
\usepackage{bm} 
\usepackage[normalem]{ulem}
\usepackage{multirow}
\usepackage{booktabs}

\usepackage[numbers]{natbib}
\usepackage{fancyhdr}

\makeatletter

\newcommand{\Rmnum}[1]{\expandafter\@slowromancap\romannumeral #1@}
\makeatother

\newcommand{\myHeaderRight}{NDT and E Intesnational}
\newcommand{\myHeaderLeft}{G. Zhang et al.}

\makeatletter
\def\ps@pprintTitle{%
     \let\@oddhead\@empty
     \let\@evenhead\@empty
     \def\@oddhead{\small\itshape \myHeaderLeft \hfill \myHeaderRight}%
     \def\@evenhead{\small\itshape \myHeaderLeft \hfill \myHeaderRight}%
     \def\@oddfoot{}%
     \def\@evenfoot{}%
}
\makeatother

\def\tsc#1{\csdef{#1}{\textsc{\lowercase{#1}}\xspace}}
\tsc{WGM}
\tsc{QE}
\tsc{EP}
\tsc{PMS}
\tsc{BEC}
\tsc{DE}

\begin{document}
\let\WriteBookmarks\relax
\def\floatpagepagefraction{1}
\def\textpagefraction{.001}

\shorttitle{Wavelet-Optimized Pseudo-3D Accelerated Diffusion Model}
\shortauthors{G. Zhang et~al.}

\title [mode = title]{Wavelet-Optimized Pseudo-3D Accelerated Diffusion Model for Truncated Computed Laminography}

\address[1]{Key Lab of Optoelectronic Technology and Systems, and Engineering Research Center of Industrial Computed Tomography Nondestructive Testing, Ministry of Education, Chongqing University, Chongqing, China, 400044}


\author[1]{Genyuan Zhang}[style=chinese]
\fnmark[1]
\ead{zhanggy@stu.cqu.edu.cn}

\author[1]{Junyao Wang}[style=chinese]
\fnmark[1]
\ead{wjy19972405100@163.com}

\author[1]{Chuandong Tan}[style=chinese]
\ead{tancd@stu.cqu.edu.cn}

\author[1]{Fenglin Liu}[style=chinese]
\cormark[1] 
\ead{liufl@cqu.edu.cn}

\author[1]{Yongning Zhou}[style=chinese]
\cormark[1] 
\ead{zynlxu@sina.com}

\cortext[cor1]{Corresponding author.}
\fntext[fn1]{These authors contributed equally to this work.}

\nonumnote{This work was supported by the National Key Research and Development Program of China (No. 2022YFF0706400), the National Natural Science Foundation of China (No. 62171067), and the Fundamental Research Funds for the Central Universities (No.2024CDJYXTD-009).}

\begin{abstract}
Computed Laminography (CL) is a key technology for the nondestructive testing of large plate-shaped objects. However, field-of-view (FOV) limitations inevitably lead to the truncation of projected data, an ill-posed inverse problem that causes severe reconstruction artifacts. Existing deep learning methods typically rely on 2D architectures that lack rigorous data consistency constraints. Furthermore, they conventionally confine artifact removal strictly to the FOV, discarding potentially recoverable information outside it. To overcome these limitations, we first introduce a comprehensive CL FOV analysis—categorizing the space into data-complete, data-incomplete, and data-free regions. By extending our reconstruction target to encompass the data-incomplete region, we significantly expand the effective imaging range and enhance scanning efficiency. To achieve this, we propose a novel wavelet-optimized pseudo-3D accelerated diffusion model for CL truncation reconstruction (CL-DM). Our method utilizes a standard 2D diffusion model for slice aggregation, combined with a 3D model-based iterative reconstruction (MBIR) method to ensure strict data consistency. To mitigate inter-slice discontinuities, we introduce wavelet regularization along the z-direction, paired with a translation-invariant (TI) mechanism and a low-frequency preservation strategy. Finally, we introduce a 3D fast sampling architecture, significantly accelerating inference speed. Extensive simulations and real-world experiments demonstrate that CL-DM is superior in effectively eliminating truncation artifacts and restoring high-fidelity, continuous 3D structures.
\end{abstract}

\begin{graphicalabstract}
\includegraphics[width=\textwidth]{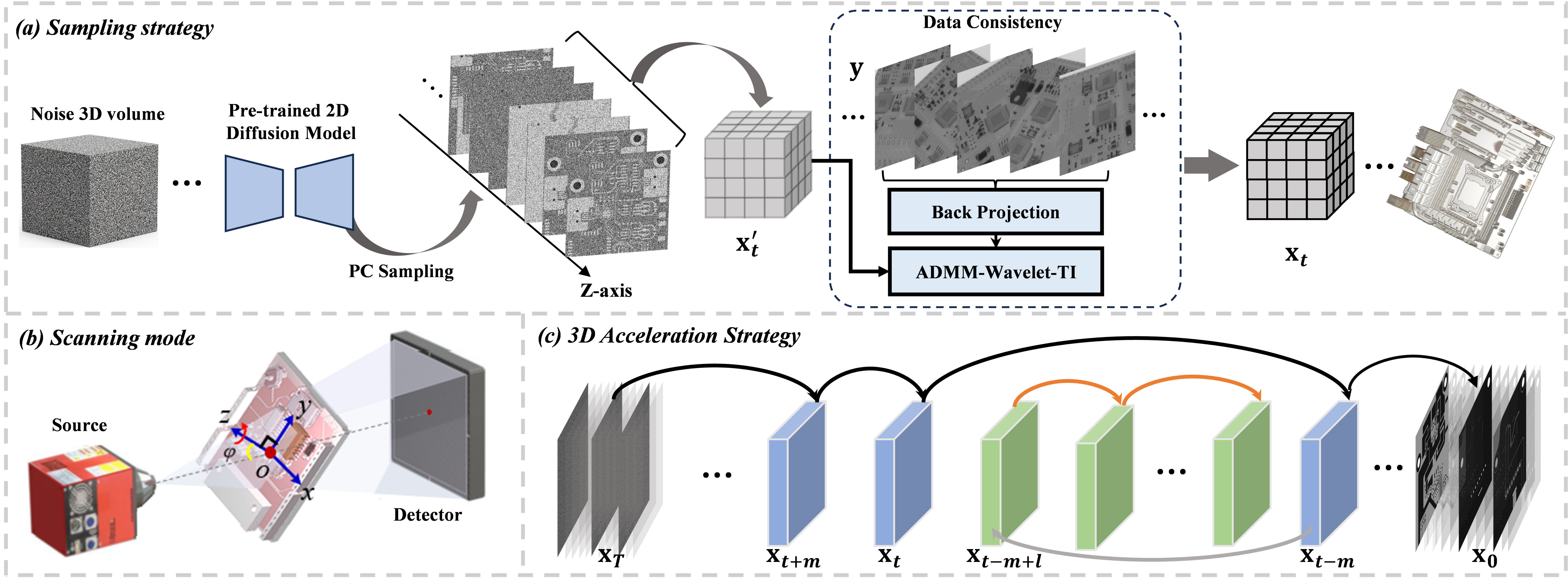}
\end{graphicalabstract}

\begin{highlights}
\item We propose a novel truncation reconstruction method that effectively recovers data-incomplete regions, expanding the effective field-of-view (FOV) and scanning efficiency.
\item We develop a highly efficient pseudo-3D diffusion strategy tailored to resolve the CL truncation problem, which strictly enforces volumetric data consistency without the computational burden of native 3D models.
\item We combine $z$-directional wavelet regularization with a translation-invariant (TI) mechanism and a low-frequency preservation strategy to robustly suppress inter-slice discontinuity artifacts and mitigate aliasing.
\item We introduce a fast-sampling architecture adapted for 3D volumetric data, successfully alleviating the bottleneck of slow inference speeds typically associated with diffusion-based generative models.
\end{highlights}

\begin{keywords}
Computed Laminography (CL) \sep accelerated diffusion models \sep wavelet-optimized
\end{keywords}

\maketitle

\section{Introduction}
Computed laminography (CL) plays an important role in the detection of plate-shaped objects with typical large aspect ratios, such as printed circuit boards (PCBs), chip packages, laminated composite materials, and paleontological fossils \cite{shi2023automatic,ghandourah2023evaluation,dorador2024computed,lu2026review}. However, due to limitations in the physical size and manufacturing cost of flat panel detectors, the field of view (FOV) of CL systems often cannot cover the entire object when performing high-resolution imaging of large plate-shaped workpieces, resulting in truncation of the projected data \cite{wood2019computed}. Since truncation does not satisfy Radon's data completeness condition, it is a typical ill-posed inverse problem. Directly using the filtered back projection (FBP) algorithm for reconstruction will produce significant cupping artifacts and attenuation value bias. Therefore, researching truncation artifact correction algorithms is of significant theoretical and engineering application value for improving the imaging quality of industrial CL systems and promoting the accurate detection of complex plate-shaped components. Traditional methods for solving CL truncation mainly include projection extrapolation, projection weighting, and regularization-based iterative methods \cite{ohnesorge2000efficient, hsieh2004novel, sourbelle2005reconstruction, tan2026truncated}. However, since the above methods rely on manual priors, the effect of truncation extrapolation is limited \cite{margosian1982redundant, cho1996cone, wang2025truncated,vogelgesang2017iterative,lu2023anisotropic}. 

Deep learning, propelled by its robust data-driven capabilities, offers promising solutions for the projection truncation problem. Depending on the domain in which the network operates, existing deep learning methods can be broadly classified into three categories \cite{huang2021data}: 1) Projection-domain methods: These approaches leverage neural networks to extrapolate or complete the missing projection data prior to analytical reconstruction \cite{lee2017view, ghani2018deep, chen2025improving}. 2) Image-domain methods: These techniques typically employ convolutional neural networks (CNNs) \cite{han2019one, Fou2019ct, gu2017multi, xie2018artifact} or generative adversarial networks (GANs) \cite{li2024pidnet, chen2024sc} to learn an end-to-end mapping from artifact-corrupted truncated reconstructions to clear images. Additionally, recent advancements utilize deep generative priors to recover the structural information lost during truncation \cite{liman2024diffusion, huang2021data}. 3) Dual-domain methods: These architectures embed physical forward models or iterative reconstruction operators directly into the deep neural network \cite{aggarwal2018modl, zhu2018image, li2019learning}.

Despite these advancements, current methodologies still face three critical limitations. Firstly, the data consistency inherent in truncated data has not been fully explored. Above methods do not impose sufficient consistency constraints on the output, leading to pseudo-structures, which are more pronounced when the projection is truncated over a large area. Secondly, for plate-like CL imaging, we are concerned not only with the effect of the cross-section but also with the effect of the layering. Above 2D-based deep learning methods often introduce discontinuity artifacts in the layering. Finally, existing methods typically focus only on artifact removal within the FOV and ignore information outside the FOV.  

Diffusion models have been widely adopted for solving inverse problems, owing to their formidable data generation capabilities and iterative refinement mechanisms \cite{daras2024survey, liu2026laminodiff}. Crucially, these models can incorporate measurement constraints during the reconstruction process, thereby strictly enforcing data consistency. Nevertheless, directly training a fully 3D diffusion model remains impractical due to prohibitive computational overhead and the scarcity of large-scale 3D industrial datasets. Given the relatively thin geometric profile of the targets typically inspected in CL imaging, leveraging a 2D diffusion model augmented with a $z$-axis Total Variation (TV) prior emerges as a highly viable and efficient alternative \cite{chung2023solving}.

In this paper, we analyzed the scanned area and found that it contained both complete and incomplete data areas. In order to recover all scanned areas, we propose a wavelet-optimized pseudo-3D accelerated diffusion model for CL truncation reconstruction (CL-DM).
First, we employ a standard 2D diffusion model and aggregate slices to achieve 3D data consistency.
Then, we introduce wavelet regularization along the z-direction to alleviate inter-slice discontinuities caused by 2D-based learning.
To reduce the additional aliasing artifacts caused by wavelet regularization, we incorporate a translation-invariant (TI) mechanism and a low-frequency preservation strategy.
Finally, We extend the 2D accelerated sampling method \cite{10570449} to 3D to balance artifact reduction and sampling speed.
This paper presents four significant contributions, which can be summarized as follows:

\begin{itemize}
    \item We propose a novel truncation reconstruction method that effectively recovers data-incomplete regions, expanding the effective FOV and scanning efficiency.
    \item We develop a highly efficient pseudo-3D diffusion strategy tailored to resolve the CL truncation problem, which strictly enforces volumetric data consistency without the computational burden of native 3D models.
    \item We combine $z$-directional wavelet regularization with a translation-invariant (TI) mechanism and a low-frequency preservation strategy to robustly suppress inter-slice discontinuity artifacts and mitigate aliasing.
    \item We introduce a fast-sampling architecture adapted for 3D volumetric data, successfully alleviating the bottleneck of slow inference speeds typically associated with diffusion-based generative models.
\end{itemize}

\begin{figure*}
\centerline{\includegraphics[width=\textwidth]{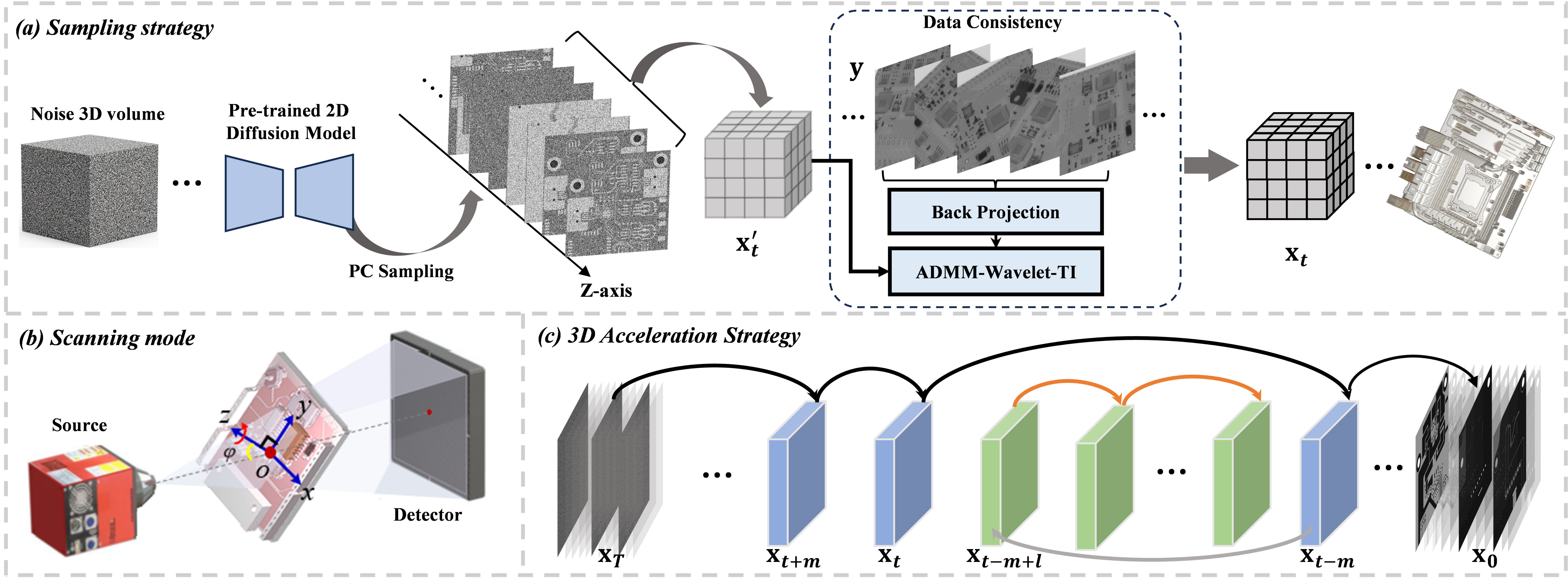}}
\caption{ Overview of the proposed CL-DM framework. (a) Slices are generated using a standard 2D diffusion model. An Alternating Direction Method of Multipliers (ADMM) is subsequently employed to enforce data consistency, incorporating a $z$-directional wavelet regularization prior. (b) Schematic diagram of the Computed Laminography (CL) scanning geometry. (c) Illustration of the 3D accelerated sampling strategy, which comprises skip sampling (black line), time-backtracking sampling (gray line), and fine sampling (yellow line).
}
\label{fig1}
\end{figure*}

\section{Related Works}
\subsection{Model-based CL Iterative Reconstruction}
In a typical industrial CL system, the physical attenuation process of X-rays passing through a plate-shaped sample can be described by a discretized set of linear equations: 
\begin{equation} 
{\bf{y}} = {\bf{Ax}} + {\bf{n}},
\label{eqr4}\end{equation}
where ${\bf{x}} \in \mathbb{R}{^{{{\rm{N}}_{\rm{z}}} \times {{\rm{N}}_{\rm{x}}} \times {{\rm{N}}_{\rm{y}}}}}$ represents the high-resolution three-dimensional voxel matrix to be reconstructed (${{{\rm{N}}_{\rm{z}}}}$ is the number of slice layers, and ${{{\rm{N}}_{\rm{x}}}}$ and ${{{\rm{N}}_{\rm{y}}}}$ are the single-layer spatial resolution); ${\bf{y}}$ is the set of multi-view two-dimensional projection observations actually acquired by the detector; ${\bf{A}}$ is the matrix characterizing the specific CL forward projection operator; and ${\bf{n}}$ is the system measurement noise that follows a Gaussian or Poisson distribution. In local high-resolution imaging modes, truncation occurs because the size of the detected target is larger than the detector's field of view (FOV). Mathematically, projection truncation causes the system to be in an underdetermined state. Therefore, the standard method for estimating the unknown image x from the truncated projection y is to perform the following regularized reconstruction: 
\begin{equation} 
{{\bf{x}}^ * } = \mathop {\arg \min }\limits_{\bf{x}} \frac{1}{2}\left\| {{\bf{y}} - {\bf{Ax}}} \right\|_2^2 + {\lambda _1}{\cal R}({\bf{x}}),
\label{eqr5}\end{equation}
where ${\cal R}$ is the appropriate regularization of ${\bf{x}}$, and ${\lambda _1}$ is the coefficient used for balancing. 

\subsection{Stochastic Differential Equation Models}
Stochastic differential equations (SDE) are the mathematical foundation of Score-based generative models (SGM) and are widely used to solve inverse problems. Let ${ {t}} \in [0,{ {T}}]$ be a continuous-time variable, ${{ {p}}_0}({\bf{x}})$ be the distribution of the real image data, and ${{ {p}}_{ {T}}}({\bf{x}}) \approx {\cal N}({\bf{0}},\sigma _{ {T}}^2{\bf{I}})$ be the distribution of pure noise. Forward SDE perturbs the data into noise, which takes the form \cite{songscore}:
\begin{equation} 
{ {d}}{\bf{x}} = f({\bf{x}},{ {t}}){ {dt}} + { {g}}({ {t}}){ {d}}{\bf{w}},
\label{eqr1}\end{equation}
where $f({\bf{x}},{ {t}})$ is the drift coefficient, ${ {g}}({ {t}})$ is the diffusion coefficient, and ${\bf{w}}$ is the standard multidimensional Wiener process. Starting from the pure Gaussian noise at time ${ {t = T}}$, the reverse integration proceeds backwards along time, ultimately generating a clear image that conforms to the original distribution. This is called the reverse SDE, and its form is as follows \cite{songscore}: 
\begin{equation} 
{ {d}}{\bf{x}} = \left[ {f({\bf{x}},{ {t}}) - { {g}}{{({ {t}})}^2}{\nabla _{{\bf{x}}_t}}\log {{ {p}}_{ {t}}}({\bf{x}})} \right]{ {dt}} + { {g}}({ {t}}){ {d}}\overline {\bf{w}},
\label{eqr2}\end{equation}
The key to obtaining a clear image through the reverse process is obtaining ${{\nabla _{{\bf{x}}_t}}\log {{ {p}}_{ {t}}}({\bf{x}})}$, which can be obtained through score matching training \cite{vincent2011connection}:
\begin{equation} 
\begin{array}{l}
{\cal L}(\theta ) = {\mathbb{E}_{{ {t}}\sim{\cal U}[0,{ {T}}],{{\bf{x}}_0}\sim{{ {p}}_{0}}({\bf{x}})}}\\
\left[ {\lambda ({ {t}})\left\| {{{\bf{s}}_\theta }({{\bf{x}}_{ {t}}},{ {t}}) - {\nabla _{{{\bf{x}}_{ {t}}}}}\log { {p}}({{\bf{x}}_{ {t}}}|{{\bf{x}}_0})} \right\|_2^2} \right]
\end{array}
\label{eqr3}
\end{equation}
where ${\lambda ({ {t}})}$ is the weighting scheme and ${{{\bf{s}}_\theta }({{\bf{x}}_{ {t}}},{ {t}})}$ is neural network.

\subsection{2D to 3D diffusion model for Solving 3D Inverse Problems}
Reconstructing 3D volumes from ill-posed inverse problems using native 3D diffusion models is fundamentally hindered by the curse of dimensionality, which imposes prohibitive memory costs and extensive data requirements. To overcome these limitations, recent generative frameworks leverage pre-trained 2D diffusion priors to enforce volumetric coherence. For instance, the DiffusionMBIR framework addresses inter-slice inconsistency by augmenting a 2D in-plane diffusion prior with a 1D total variation (TV) regularizer along the longitudinal axis \cite{chung2023solving}. Because simple mathematical regularizers may fail to fully capture global structural dependencies , the Two Perpendicular 2D Diffusion Models (TPDM) approach represents the 3D data distribution as a product of orthogonal 2D constituents \cite{lee2023improving}. This allows the model to capture global dependencies without relying on hand-crafted penalties. Alternatively, to optimize computational efficiency, the Two-and-a-half Order Score-based Model (TOSM) utilizes a single pre-trained 2D network to compute weighted pseudo-3D scores across three orthogonal planes \cite{li2024two}.

\section{Method}
\subsection{CL Imaging FOV Analysis}
The projection truncation phenomenon, which is ubiquitous in practical inspections due to detector size limitations, can destroy sampling completeness and induce characteristic artifacts. 

\begin{figure}[pos=t]
\centerline{\includegraphics[width=\linewidth]{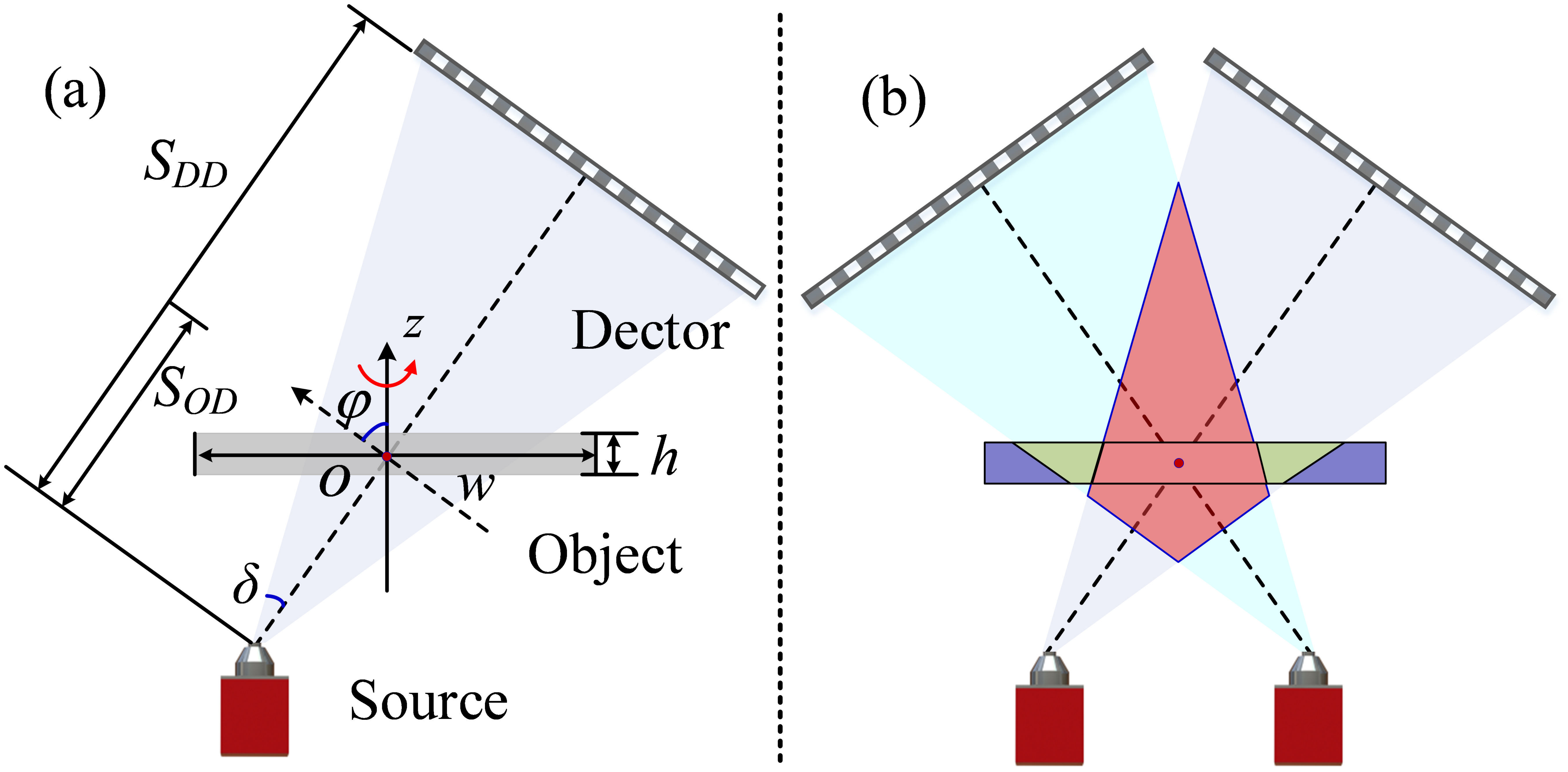}}
\caption{Schematic diagram of sampling region division in truncated scanning mode. (a) RCL scanning mode; (b) Sampling region division.}
\label{sacn}
\end{figure}

As shown in Figure \ref{sacn} (a), in the CL scanning mode, the object rotates with the platform at a rotation angle $\theta$. Here, $S_{DD}$ is the distance from the X-ray source to the detector, $S_{OD}$ is the distance from the X-ray source to the object's rotation center, $m$ is the total number of detector elements, $d$ is the detector pixel width, and $\varphi$ is the CL scanning tilt angle. Assuming the reconstructed voxel matrix dimensions of the scanned object are $l \times w \times h$ (with $h$ in the thickness direction) and the voxel size is $d'$, its maximum physical cross-sectional radius is defined as $R_{max} = \frac{d^{\prime}}{2}\sqrt{l^{2}+w^{2}}$. The origin $O(0,0,0)$ is set at the geometric center of the inspected PCB. Let the physical coordinates of an arbitrary voxel point within the PCB be $(x,y,z)$; its polar coordinate representation is $(r,\theta,z)$, where $r = \sqrt{x^{2}+y^{2}}$. Based on the spatial domain analysis of the CL scanning mode, the sampling area can be divided into three distinct regions: the Complete Sampling Region, the Incomplete Sampling Region, and the Unsampled Region.

The mathematical definitions of three regions are as follows (detailed derivations are provided in Appendix \ref{a}):

\begin{enumerate}
    \item \textbf{Complete Sampling Region}: Defined as the spatial intersection of all beams when the source rotates $360^\circ$. Its theoretical set $\Omega_{CSR}$ is strictly defined as:
    \begin{equation}
        \begin{aligned}
            \Omega_{CSR} = \Big\{ &(r,\theta,z) \;\Big|\; r \le \min(R_{up}(z), R_{dn}(z),  \\
            R_{trans}(z)), & z \in \left[-\frac{h \cdot d^{\prime}}{2}, \frac{h \cdot d^{\prime}}{2}\right] \Big\}
        \end{aligned}
    \end{equation}

    \item \textbf{Incomplete Sampling Region}: Areas penetrated by rays only at partial scanning angles. Its mathematical set $\Omega_{incomp}$ is defined as:
    \begin{equation}
        \begin{aligned}
        \Omega_{incomp} = \Big\{ &(r,\theta,z) \;\Big|\; \min(R_{up}, R_{dn}, R_{trans}) \le r \\
         & \le \min(R_{sweep}, R_{max}), \\
        & z \in \left[-\frac{h \cdot d^{\prime}}{2}, \frac{h \cdot d^{\prime}}{2}\right] \Big\}
        \end{aligned}
    \end{equation}

    \item \textbf{Unsampled Region}: Areas that are not penetrated by rays at any angle, typically manifesting as information voids. Its mathematical set $\Omega_{blind}$ is:
    \begin{equation}
        \begin{aligned}
            \Omega_{blind} = \Big\{ &(r,\theta,z) \;\Big|\; R_{sweep}(z) \le r \le R_{max}, \\
            & z \in \left[-\frac{h \cdot d^{\prime}}{2}, \frac{h \cdot d^{\prime}}{2}\right] \Big\}
        \end{aligned}
    \end{equation}
\end{enumerate}

In practical PCB inspections, these regions are distributed as shown in Figure \ref{sacn} (b), where the red,  green, and blue-purple areas represent the complete sampling, incomplete, and unsampled regions, respectively. In the reconstructed image, the area inside the truncated ring represents the region with complete data, while the area outside the truncated ring represents other regions. Since the recovery results from unsampled regions are unreliable, we will only consider recovering regions with incomplete data.

\subsection{Pseudo-3D Strategy}
Directly training a diffusion model for processing 3D data faces catastrophic memory overhead and insurmountable computational burdens, making it impractical for industrial applications. On one hand, considering that the objects detected by CL are plate-like structures with thin z-direction thickness; on the other hand, plate-like structures contain the most features within their horizontal cross-sections. Therefore, we employ 2D priors combined with z-direction regularization to mitigate the interlayer discontinuity problem in directly applying a 2D diffusion model to the CL truncation problem.

\noindent {\bf{2D Diffusion Priors}}
In the inverse SDE solution, a Predictor-Corrector (PC) architecture is adopted, and its discretization form is as follows \cite{songscore}:
\begin{equation} 
{{{\bf{x'}}}_{{i}}} \leftarrow {{\bf{x}}_{{{i}} + 1}} + (\sigma _{{{i}} + 1}^2 - \sigma _{{i}}^2){\bf{s}}_\theta ^ * ({{\bf{x}}_{{{i}} + 1}},\sigma_{{i} + 1}) + \sqrt {\sigma _{{{i}} + 1}^2 - \sigma _{{i}}^2} \epsilon,
\label{eqm1}\end{equation}
where ${\bf{s}}_\theta ^ *$ represents the learned 2D score. By updating the 3D volume layer by layer using the above formula, rich 2D prior information can be obtained.

\noindent {\bf{Wavelet-based Regularization in the z-direction}}
After all slices undergo one round of PC sampling according to Eq. \ref{eqm1}, we introduce 3D data consistency to ensure that the generated image sequence conforms to the original data distribution:
\begin{equation} 
{{\bf{x}}^ * } = \mathop {\arg \min }\limits_{\bf{x}} \frac{1}{2}\left\| {{\bf{y}} - {\bf{Ax}}} \right\|_2^2 + \lambda {\left\| {{\Psi _{\rm{z}}}({\bf{x}})} \right\|_1},
\label{eqm2}\end{equation}
where ${\bf{A}}$ is the 3D CL truncation projection operator, ${\Psi _{\rm{z}}}( \cdot )$ represents the Haar wavelet transform along the z-axis. The priors for the xy plane are obtained through the neural network ${\bf{s}}_\theta ^ * $. We only utilize the sparsity of the wavelet domain to remove the z-axis artifacts and maintain interlayer continuity. Unlike TV regularization \cite{chung2023solving}, wavelet regularization can not only more precisely characterize the sparsity of high-frequency features, but more importantly, the Haar wavelet basis has strict orthogonality, a mathematical property that simplifies the subsequent solution structure \cite{fan2015multi}. In addition, Figure \ref{figTI} shows that the background noise is higher with TV regularization than with wavelet regularization.

\noindent {\bf{Algorithm solution}}
Since the $L_1$ norm in the wavelet domain is non-smooth and non-differentiable, directly differentiating the objective function cannot yield a closed-form solution. Therefore, we use Alternating Direction Method of Multipliers (ADMM) \cite{neal2011distributed} for efficient decoupling. By introducing auxiliary variables ${{\bf{z}}_{{wave}}} = {\Psi _{z}}({\bf{x}})$ from the wavelet domain, we transform the original unconstrained optimization into an equality-constrained optimization:
\begin{equation} 
{\min _{{\bf{x}},{{\bf{z}}_{{\rm{wave}}}}}}\frac{1}{2}\left\| {{\bf{y}} - {\bf{Ax}}} \right\|_2^2 + \lambda {\left\| {{{\bf{z}}_{{\rm{wave}}}}} \right\|_1}{{ s}}{{.t}}{{.}}\quad {{\bf{z}}_{{\rm{wave}}}} = {\Psi _{\rm{z}}}({\bf{x}}).
\label{eqm3}\end{equation}
Constructing an augmented Lagrangian function in scale form:
\begin{equation} 
\begin{array}{l}
{{\cal L}_\rho }({\bf{x}},{{\bf{z}}_{{wave}}},{{\bf{u}}_{{wave}}}) = \frac{1}{2}\left\| {{\bf{y}} - {\bf{Ax}}} \right\|_2^2 + \lambda {\left\| {{{\bf{z}}_{{wave}}}} \right\|_1}\\
 + \frac{\rho }{2}\left\| {{\Psi _{z}}({\bf{x}}) - {{\bf{z}}_{{wave}}} + {{\bf{u}}_{{wave}}}} \right\|_2^2
\end{array},
\label{eqm4}\end{equation}
where ${{\bf{u}}_{{wave}}}$ is the scaled dual variable (i.e., the Lagrange multiplier, used to penalize the degree of constraint non-satisfaction during iteration, i.e., the error of ${\Psi _{z}}({\bf{x}}) - {\bf{z}}$), and $\rho  > 0$ is the penalty parameter controlling the strength of the quadratic penalty. The ADMM algorithm updates ${\bf{x}}$, ${{\bf{z}}_{{wave}}}$, and ${{\bf{u}}_{{wave}}}$ by sequentially fixing the remaining variables and alternately minimizing them. The solution to the problem is shown in Appendix \ref{b}.

\subsection{Artifact Suppression Based on Translation Invariance}
\begin{figure}[pos=t]
\centerline{\includegraphics[width=\linewidth]{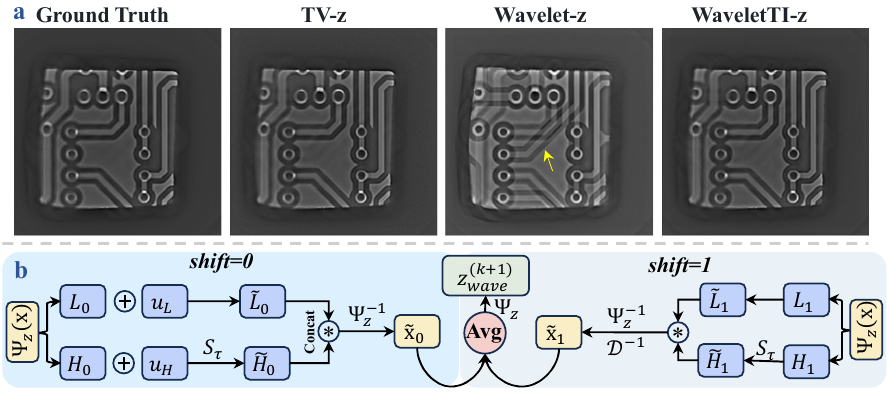}}
\caption{a: The comparison of different regularization methods, from left to right, is ground truth, TV regularization in the z-direction, wavelet regularization in the z-direction, and wavelet regularization in the z-direction with TI. b: A schematic diagram of the TI mechanism and the low-frequency protection strategy.}
\label{figTI}
\end{figure}

The standard Haar Discrete Wavelet Transform (DWT) regularization term can easily exacerbate aliasing artifacts in CL reconstruction as shown in Figure \ref{figTI} (a). To address this, we propose a Translation Invariant (TI) mechanism and a low-frequency protection strategy.

The 3D image is cyclically roll along the z-axis in several steps. Wavelet decomposition and soft thresholding are performed at different roll phases, followed by inverse roll to align them. Finally, the arithmetic mean is taken in the image domain. Specifically, let the roll operator be denoted ${{\cal D}_{{ \text{shift}}}}( \cdot )$. We choose $ \text{shift}=0$ and $ \text{shift}=1$ (${{\cal D}_{{ \text{shift}}}}( \cdot )$). For the $ \text{shift}=0$, we extract coefficients ${{c}_0} = {\Psi _{z}}(\bf{x})$ from the forward wavelet transform, and then strictly divide the wavelet coefficients into low-frequency approximation components ${{L}_0} = {{c}_0}[0:{{N}_{z}}/2]$ and high-frequency detail components ${{H}_0} = {{c}_0}[{{N}_{z}}/2:{{N}_{z}}]$. The corresponding dual variables are also split: ${{u}_{L}}$ and ${{u}_{H}}$. Then, a differentiated thresholding strategy is implemented: for ${{L}_0}$, no nonlinear shrinkage is applied, only pure ADMM linear variable recombination is performed:
\begin{equation} 
{{{\tilde L}}_0} = {{{L}}_0} + {{{u}}_{{L}}}.
\label{eqm5}\end{equation}
For the high-frequency component ${{H}_0}$, a soft shrinkage operator with a threshold of ${{\cal S}_\tau }$ is applied:
\begin{equation} 
{{{{\tilde H}}}_0} = {{\cal S}_\tau }({{{H}}_0} + {{{u}}_{{H}}}).
\label{eqm6}\end{equation}
After recombination, the denoised coefficients are obtained, that is ${{{{\tilde c}}}_0} ={\mathop{\rm Concat}\nolimits}  [{{{{\tilde L}}}_0},{{{{\tilde H}}}_0}]$, which are then restored to the image domain by inverse Haar transform to obtain ${{\tilde x}_0} = \Psi _{{z}}^{ - 1}({{{{\tilde c}}}_0})$. For the $\text{shift}=1$ channel, considering its main function is to smooth boundary artifacts, to ensure algorithm stability and simplify the maintenance of the dual variables, we update it as follows: \begin{equation} 
{{{{\tilde L}}}_1}{{ = }}{{{L}}_1},{{{{\tilde H}}}_1} = {{\cal S}_\tau }({{{H}}_1}).
\label{eqm7}\end{equation} 
After recombination, an inverse wavelet transform is performed, followed by inverse rolling reset: 
\begin{equation} 
{{\tilde {\bf{x}}}_1} = {{\cal D}_{ - 1}}\left( {\Psi _{{{\bf{z}}}}^{ - 1}({\mathop{\rm Concat}\nolimits} [{{{{\tilde L}}}_1},{{{{\tilde H}}}_1}])} \right).
\label{eqm8}\end{equation} 
Finally, the denoising results of the two phases are fused by arithmetic mean in the physical image domain and then projected forward back to the wavelet reference domain to complete the update of the z variable: 
\begin{equation} 
{{\bf{x}}_{{{avg}}}} = \frac{1}{2}({{\tilde {\bf{x}}}_0} + {{\tilde {\bf{x}}}_1}),{\bf{z}}_{{{wave}}}^{({{k}} + 1)} = {\Psi _{{z}}}({{\bf{x}}_{{{avg}}}}).
\label{eqm9}\end{equation} 
The overall calculation diagram is shown in Figure \ref{figTI} (b).

\subsection{3D Acceleration Strategy}
In score-based generative models, standard reverse sampling for 3D reconstruction incurs prohibitive computational costs. While simplistic jump-sampling accelerates inference, the aggressively expanded noise scale severely degrades high-frequency details. To achieve a dynamic balance between efficiency and reconstruction fidelity, we adopt the Time-Reversion Fast-Sampling (TIFA) framework \cite{10570449} and introduce two critical adaptations for high-dimensional volumetric data: a. We generalize the jump-sampling and time-reversion mechanisms to 3D volumes, allowing macroscopic structure constraints to form rapidly under limited computational budgets. b. We replace the original Diagonal Total Variation (DTV) prior in TIFA with a 3D ADMM-Wavelet operator ($\mathcal{H}_{ADMM}$). 

Given a total of $T$ discrete diffusion steps, we construct a sparse sampling trajectory with $N$ steps ($N \ll T$). The method executes coarse-grained jump-sampling with a span of $m = T/N$, followed by fine-grained time-reversion over $L$ steps. During the re-sampling phase, intermediate states are refined using a Predictor-Corrector (PC) sampler based on the Variance Exploding SDE (VE-SDE). For an intermediate state $x_{t+l}^{\prime}$ at time step $t+l$, the predictor performs a numerical SDE update:
\begin{equation} 
\begin{array}{l}
{\bf{x}}_{t + l + 2/3}^\prime  = {\bf{x}}_{t + l + 1}^\prime  + (\sigma _{t + l + 1}^2 - \sigma _{t + l}^2)s_\theta ^*({\bf{x}}_{t + l + 1}^\prime ,t + l + 1)\\
 + \sqrt {\sigma _{t + l + 1}^2 - \sigma _{t + l}^2} {\bf{z}}
\end{array}
\label{eqm10}\end{equation} 
The corrector then applies Langevin dynamics for state refinement:
\begin{equation} 
{\bf{x}}_{t+l+1/3}^{\prime} = {\bf{x}}_{t+l+2/3}^{\prime} + \epsilon_{t+l} s_{\theta}^{*}({\bf{x}}_{t+l+2/3}^{\prime}, t+l) + \sqrt{2\epsilon_{t+l}} {\bf{z}}
\label{eqm11}\end{equation} 
Following the PC update, we apply our proposed 3D ADMM-Wavelet operator using the measurement data $y$ to enforce data consistency and structural correction:
\begin{equation} 
{\bf{x}}_{t+l}^{\prime\prime} = \mathcal{H}_{ADMM}({\bf{x}}_{t+l+1/3}^{\prime}, {\bf{y}}).
\label{eqm12}\end{equation} 
By decoupling the coarse structural generation and fine-grained detail recovery, stable 3D reconstruction is guaranteed. Effective acceleration is achieved provided the jump span exceeds the time-reversion overhead, simplified by the condition $m - 1 > L$.

\section{Experiments}
\subsection{Datasets}
\begin{figure}[pos=t]
\centerline{\includegraphics[width=\linewidth]{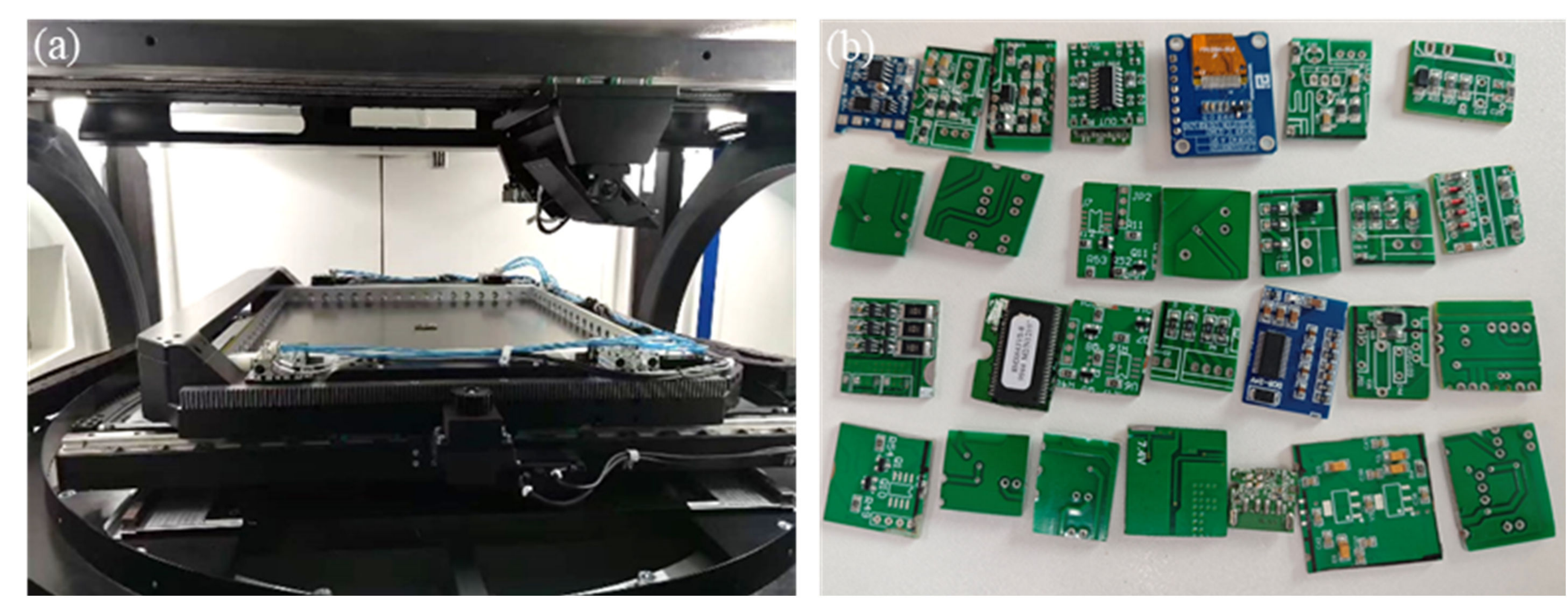}}
\caption{real world CL imaging: (a) CL system; (b) Scanned object}
\label{figsim_data}
\end{figure}

To acquire non-truncated projection data for ground truth in our simulation experiments, large circuit boards were sectioned into smaller coupons to conform to the detector's field of view. Subsequently, truncated reconstruction results were generated via simulation. To ensure data diversity and verify the algorithm's robustness under different texture features, this study selected different batches of double-layer circuit boards with varying circuit layouts as experimental subjects. The laboratory CD-700BX/µCL device was used, and the scanning system and representative samples of the real world data are shown in Figure \ref{figsim_data}. 
\subsubsection{Simulation Datasets}
In the creation of the simulation experiment dataset, for each batch of samples, they were destructively cut into small pieces that fit the field of view. High-fidelity images were then reconstructed using the CL-FBP algorithm to obtain the baseline ground truth, and two types of dedicated datasets were constructed based on this: a non-global truncated dataset and a global truncated dataset. Each dataset was randomly divided into a training set (80\%), a validation set (10\%), and a test set (10\%). The datasets was normalized to a 0-1 float32 format.

{\bf{Non-global Truncated datasets:}} In the non-global truncated case, projection truncation occurs only at certain view angles. This study selected 25 double-layer circuit boards, each with 16–25 layers cut, totaling 517 original slices. Data augmentation was performed by rotating the original slices by 90° and horizontally flipping them, ultimately constructing a dataset containing 1551 slices, each image measuring 256×256 pixels. After obtaining the baseline ground truth, non-global truncated scanning was simulated using the scanning parameters shown in Table \ref{scanning_parameters} to obtain the corresponding low-quality truncation images, thus constructing the dataset.

{\bf{Global Truncated Datasets:}} The global truncated scenario corresponds to the more severe concavity problem in industrial inspection, where the region of interest is completely surrounded by an object, and projection data from all angles are truncated. In this scenario, multilayer server motherboards were selected as the inspection object. Due to the large size of the original circuit board, it must be destructively cut into smaller pieces to fit the field of view. Local high-fidelity images were then reconstructed using the CL-FBP algorithm to obtain local ground truth images. The original large board was divided into 25 modules, each containing 40 slices, for a total of 1000 slices. To fully expand the data, an eight-angle rotation enhancement strategy was adopted, and the stacked structure was combined without omission, ultimately obtaining 8000 ground truth images, each with a size of 256×256. After obtaining the ground truth, based on the scanning parameters shown in Table \ref{scanning_parameters}, a global truncated scan was simulated through digital simulation to obtain the corresponding low-quality truncation images, thus constructing the dataset.

\subsubsection{Real World Datasets}
To verify the effectiveness of the method in real-world experiments, we conducted a real-world truncation experiment. In this experiment, the obtained projection data was itself truncated. This truncation was not obtained through simulation but directly through the device, thus conforming to reality. Corresponding to the simulation experiment, we conducted two experiments: one with non-global truncation and the other with global truncation. The experimental parameters are shown in the Table \ref{scanning_parameters}. 


\begin{table*}[t]
\centering
\caption{Simulation and real world experimental parameters for non-global truncation versus global truncation}
\label{scanning_parameters}
\begin{tabular}{lcclc}
\toprule
\multicolumn{2}{c}{Non-global truncated parameters} & & \multicolumn{2}{c}{Global truncated parameters} \\
\cmidrule(r){1-2} \cmidrule(l){4-5}
Parameter & \begin{tabular}[c]{@{}c@{}}Simulation \\ (real world)\end{tabular} & & Parameter & \begin{tabular}[c]{@{}c@{}}Simulation \\ (real world)\end{tabular} \\
\midrule
\begin{tabular}[c]{@{}l@{}}Source-to-object distance\\($S_{OD}$) / mm\end{tabular} & 152.1 & & \begin{tabular}[c]{@{}l@{}}Source-to-object distance\\($S_{OD}$) / mm\end{tabular} & 56.0 (74.7) \\
\addlinespace
\begin{tabular}[c]{@{}l@{}}Source-to-detector distance\\($S_{DD}$) / mm\end{tabular} & 609.8 & & \begin{tabular}[c]{@{}l@{}}Source-to-detector distance\\($S_{DD}$) / mm\end{tabular} & 513.4 (528.9) \\
\addlinespace
Image matrix / pixel & $256 \times 256 \times 16$ (20) & & Image matrix / pixel & \begin{tabular}[c]{@{}c@{}}$256 \times 256 \times 20$  \\ ($256 \times 256 \times 40$)\end{tabular} \\
\addlinespace
Detector matrix / pixel & $224 \times 224$ & & Detector matrix / pixel & $384 \times 384$  \\
\addlinespace
Pixel size / $\text{mm}^2$ & $0.4 \times 0.4$ & & Pixel size / $\text{mm}^2$ & $0.4 \times 0.4$  \\
\addlinespace
Tilt angle $\phi$ / $^\circ$ & 50 & & Tilt angle $\phi$ / $^\circ$ & 50 \\
\addlinespace
\begin{tabular}[c]{@{}l@{}}Number of sampling\\angles\end{tabular} & 512 & & \begin{tabular}[c]{@{}l@{}}Number of sampling\\angles\end{tabular} & 512 \\
\bottomrule
\end{tabular}
\end{table*}

\subsection{Implementation Details}
Our experiments were conducted on a server equipped with Pytorch. The training process involved 280 epochs, and a batch size was set to 2. We did not perform any checkpoints selection on the model and only selected the latest checkpoints. The entire training process took approximately 35 hours. The training procedure of VE-SDE was implemented by strictly following the guidelines recommended by Song et al. \cite{songscore}, e.g. $\delta t \in [\delta_{\text{min}}, \delta_{\text{max}}] = [0.01, 378]$, and the learning rate $lr = 2 \times 10^{-4}$. Unless otherwise specified, we set hyperparameters $\rho $ and $\lambda $ to 23.5 and 0.185, respectively. The sampling steps  $T$ was set to 16, and the backtracking step $L$ was set to 8. Both forward and backward projection operations in the reconstruction process were implemented using the CUDA kernel of ASTRA Toolbox \cite{palenstijn2013astra}. The experimental hardware platform was configured as follows: CPU: 13th Gen Intel(R) Core(TM) i5-13600KF, GPU: NVIDIA GeForce RTX 4090D, 24GB video memory. 

To comprehensively evaluate the algorithm performance, this study designed two types of comparative experiments.
The first type was a traditional physics-driven algorithm based on the imaging geometry model. First, the CL analytical FBP was selected as the basic analytical benchmark. Second, TS‑FBP \cite{wang2025truncated} was selected, and geometric weighting was introduced into the projection to reduce artifacts. Furthermore, the simultaneous iterative reconstruction technique (SIRT) was selected as a representative of algebraic reconstruction methods, approximating the solution by minimizing the reprojection error. In this study, the number of iterations was set to 700 to ensure sufficient convergence.
The second type was a deep learning data-driven algorithm that utilized neural networks to learn complex reconstruction mappings. The classic FBPConvNet \cite{jin2017deep} was selected as a representative discriminative model for end‑to‑end artifact removal. The training was conducted for 2000 epochs with a batch size of 8, using the Adam optimizer and a learning rate of 0.0002. Finally, image restoration with mean-reverting stochastic differential equations (IR‑SDE) \cite{luo2023image} was introduced as a representative generative model. Its total number of training iterations was set to 700,000, with an initial learning rate of $10^{-4}$. The MultiStepLR strategy was employed to halve the learning rate at 200,000, 400,000, and 600,000 iterations. The time step was set to $T=100$, the maximum noise scale was $\sigma_{\max}=10$, and the cosine noise schedule was adopted.

We conducted subjective evaluations at both the x-y and y-z slices. Subjective evaluations were based on visual inspection, assessing the effectiveness of truncation artifact removal and detail preservation in images generated by different methods. Objective evaluations employed commonly used peak signal-to-noise ratio (PSNR) and structural similarity index (SSIM), evaluating and calculating them on a 3D volumetric basis. Gradient information is very important in CL reconstruction. To measure the degree of gradient distortion of various algorithms, we introduced the Gradient Magnitude Similarity Deviation (GMSD) index \cite{xue2013gradient}. The larger the index, the greater the difference between the gradient and the true value.

\subsection{Performance Comparison on Simulated Datasets}
In this section, we evaluated the performance of different models on non-global truncated datasets and global truncated datasets.
\subsubsection{Evaluation on Non-global Truncated Datasets}
\begin{figure*}[pos=t]
\centerline{\includegraphics[width=\linewidth]{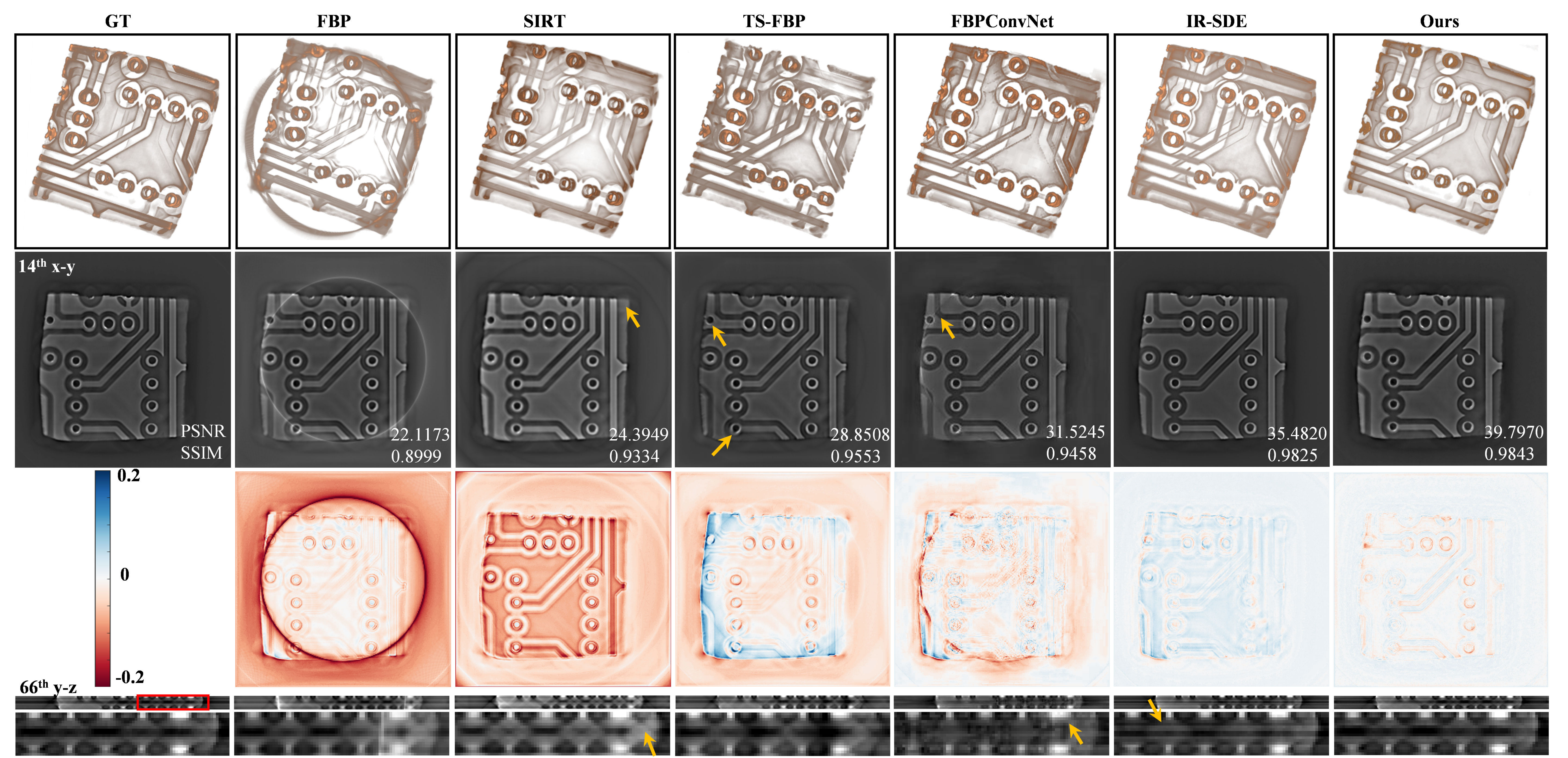}}
\caption{Qualitative results of non-global truncated data. From top to bottom: 3D rendering results, in-plane results, difference map, and layered results. The window for the in-plane results is [0,1], and the window for the layered results is [0.1,0.7]. The PSNR and SSIM indices are shown in the lower right corner of the in-plane map. From left to right: ground truth, FBP algorithm reconstruction results, and various comparative algorithms; the last column shows the proposed algorithm.}
\label{ng}
\end{figure*}

Figure \ref{ng} shows representative results of removing non-global truncated artifacts using different methods for visual comparison. The FBP results show obvious truncation artifacts, manifested in truncation bright ring artifacts and deviations in the overall attenuation value. SIRT is a commonly used iterative strategy for removing CT truncation artifacts, effectively removing truncation bright rings. However, its reconstructed images have two significant drawbacks: grayscale offset in the background region and excessive smoothing of structures, as shown by the yellow arrows. The possible reasons are underdetermined projection equations at the truncation boundaries during the iteration process. TS-FBP, as an analytical reconstruction algorithm specifically designed for CL truncation artifacts, can effectively remove truncation artifacts. However, as shown by the yellow arrow in the figure, its ability to recover the structure outside the truncated region is weak. This may be because, as an analytical method, it does not perform well on the projection of the supplementary truncation, thus resulting in poor performance outside the truncation region. In contrast, FBPConvNet, a classic end-to-end deep neural network used in CT reconstruction, outperforms analytical and iterative reconstruction methods in recovering overall attenuation information, as can be seen from the difference map. However, its ability to eliminate truncated bright ring artifacts is insufficient, as indicated by the yellow arrow. This may be due to its limited generalization ability. In contrast, the diffusion model-based method IR-SDE can effectively remove artifacts from the in-plane, but it can also cause structural discontinuities in the strata, which is due to its learning of the 2D data distribution. By incorporating regularization along the z-direction, our method achieved improved inter-slice consistency and also substantially outperformed IR-SDE in the in-plane results.

Table \ref {tab:reconstruction_results} presents the quantitative results of all methods. Our method significantly outperforms the others. Specifically, compared to the second-best IR-SDE, our method achieved approximately +8.54 dB, +2\% SSIM, and -0.0178 GMSD. In addition, the grayscale curves in Figure \ref{gray} also show that our method effectively suppresses truncation artifacts and achieves a grayscale distribution that is closest to the ground truth.

\begin{table*}[h]
\centering
\caption{Comparison of Reconstruction Methods under Lateral and Global Truncated}
\label{tab:reconstruction_results}
\begin{tabular}{ccccccc}
\toprule
\multirow{2.5}{*}{Methods} & \multicolumn{3}{c}{Non-global truncated} & \multicolumn{3}{c}{Global truncation} \\
\cmidrule(lr){2-4} \cmidrule(lr){5-7}
 & PSNR(dB) & SSIM & GMSD & PSNR(dB) & SSIM & GMSD \\
\midrule
FBP        & 22.81 & 0.89 & 0.11 & 23.45 & 0.72 & 0.19 \\
SIRT       & 24.23 & 0.92 & 0.07 & 22.81 & 0.71 & 0.19 \\
TS-FBP     & 26.10 & 0.93 & 0.08 & 25.12 & 0.79 & 0.16 \\
FBPConvNet & 27.53 & 0.91 & 0.07 & 25.01 & 0.73 & 0.16 \\
IR-SDE     & 31.82 & 0.96 & 0.04 & 25.03 & 0.77 & 0.14 \\
\textbf{Ours} & \textbf{39.78} & \textbf{0.98} & \textbf{0.02} & \textbf{28.99} & \textbf{0.83} & \textbf{0.09} \\
\bottomrule
\end{tabular}
\end{table*}

\subsubsection{Evaluation on Global Truncated Datasets}     
\begin{figure*}[pos=t]
\centerline{\includegraphics[width=\linewidth]{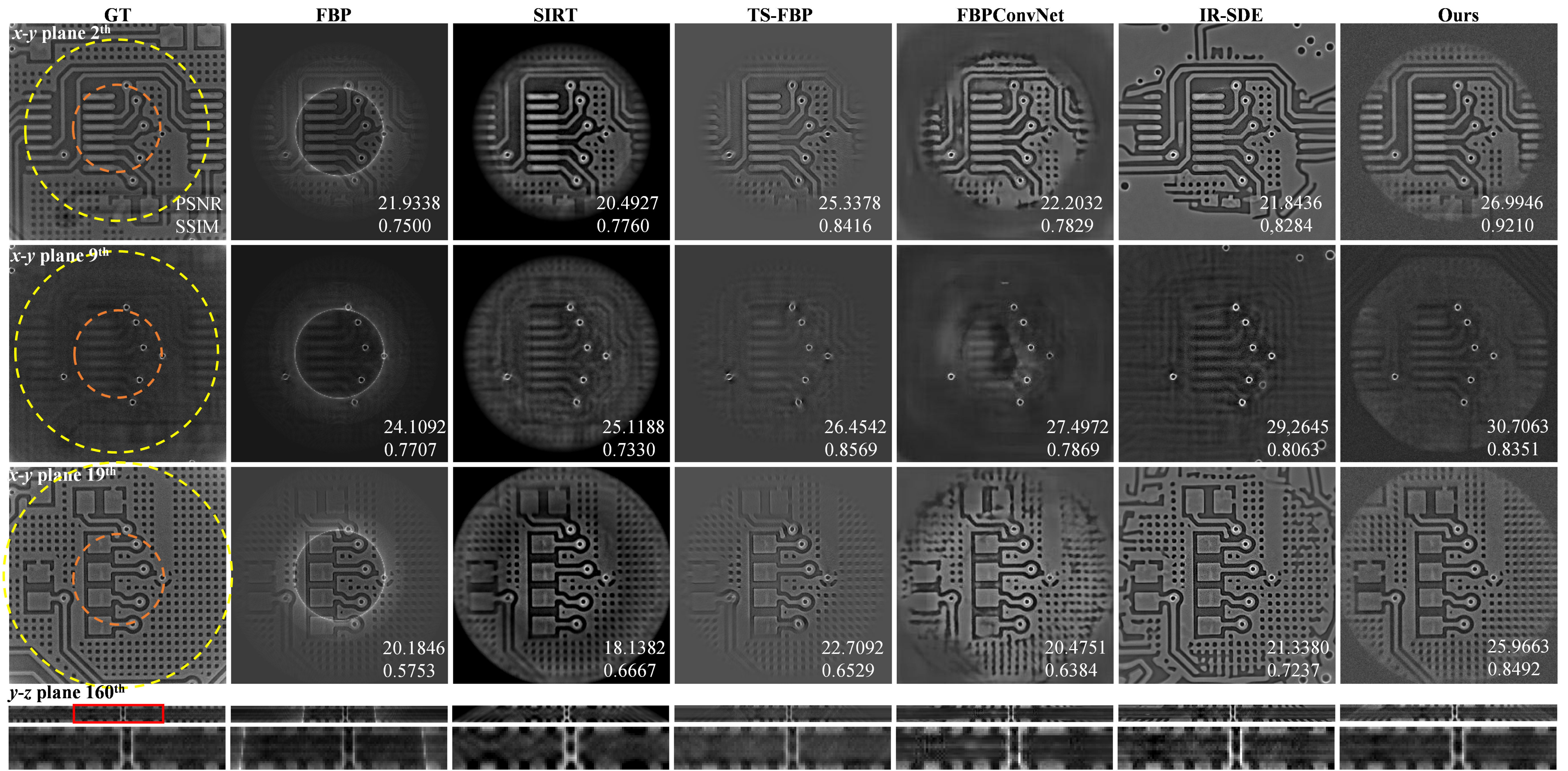}}
\caption{Qualitative results of global truncated data. In GT, the area outside the yellow dashed circle is the unsampled area, the area inside the orange circle is the data complete area, and the area between two circles is the incomplete data area. For the results of all methods, we only retained the area inside the circle. From top to bottom: three in-plane results and layered results. The window for the in-plane results is [0,1], and the window for the layered results is [0.1,0.7]. The PSNR and SSIM indices are shown in the lower right corner of the in-plane map. From left to right: ground truth, FBP algorithm reconstruction results, and various comparative algorithms; the last column shows the proposed algorithm.}
\label{g}
\end{figure*}

The Figure \ref{g} presents the qualitative results of global truncation. Global truncation is more difficult because it involves a larger truncation range and more missing data. In cases of internal truncation, all data outside the yellow dashed circle is missing, such as our CL imaging FOV analysis. Therefore, we only compared the performance of various algorithms within the dashed circle. Based on the reconstruction results of FBP, the diameter of the truncated bright ring are much smaller than those under non-global truncation. The data outside the truncation bright rings show obvious abnormal attenuation values and unclear structure. The SIRT method aims to mitigate artifacts caused by truncated bright rings, but it significantly exacerbates aliasing, as shown by the yellow arrow in the figure. This is due to insufficient prior knowledge. The TS-FBP method results in poor structural recovery in the truncated outer region. FBPConvNet suffers from noticeable artificial artifacts due to its limited generalization ability, as shown by the yellow arrows. The diffusion-based method is significantly superior to the methods mentioned above due to its powerful data modeling capabilities. However, structurally flawed regions can still be observed in the IR-SDE method, possibly due to its lack of methods for maintaining data consistency. In contrast, our method incorporates a model-based iterative approach to enforce consistency between the generated results and the measured data, thus mitigating the generation of artificial structures.

\begin{figure*}[pos=t]
\centerline{\includegraphics[width=0.9\linewidth]{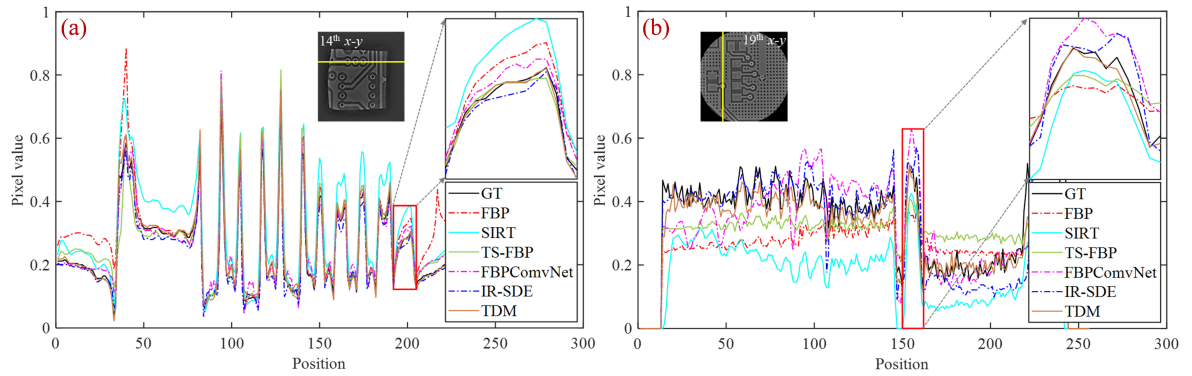}}
\caption{Gray-scale distribution of different methods on the yellow line. (a) Non-global truncation, (b) Global truncation.}
\label{gray}
\end{figure*}

Table \ref {tab:reconstruction_results} presents the quantitative results of all methods. We evaluated only the 3D objective metrics within the yellow dashed circle. We found that, except for the FBP algorithm, the metrics of all other algorithms degraded compared with the non-global truncation scheme, which was caused by severe data loss. However, our method remains optimal. Moreover, the grayscale curves in Figure \ref{gray} also show that our method effectively suppresses truncation artifacts and achieves a grayscale distribution that is closest to the ground truth.

\subsection{Performance Comparison on Real-world Datasets}  
The difference between real-world and simulation experiments is that our truncated projections are obtained through scanning with real equipment, not through simulation. Similarly, two scans were performed: one with lateral truncation and one with global truncation, yielding the respective truncated projection datasets. Furthermore, for all real-world experiments, we did not retrain the model; we directly used the model trained in the simulation experiments to generalize in the real world. Therefore, this experiment also tests the model's generalization ability.
\subsubsection{Evaluation on non-global Truncated Datasets}
\begin{figure*}[pos=t]
\centerline{\includegraphics[width=\linewidth]{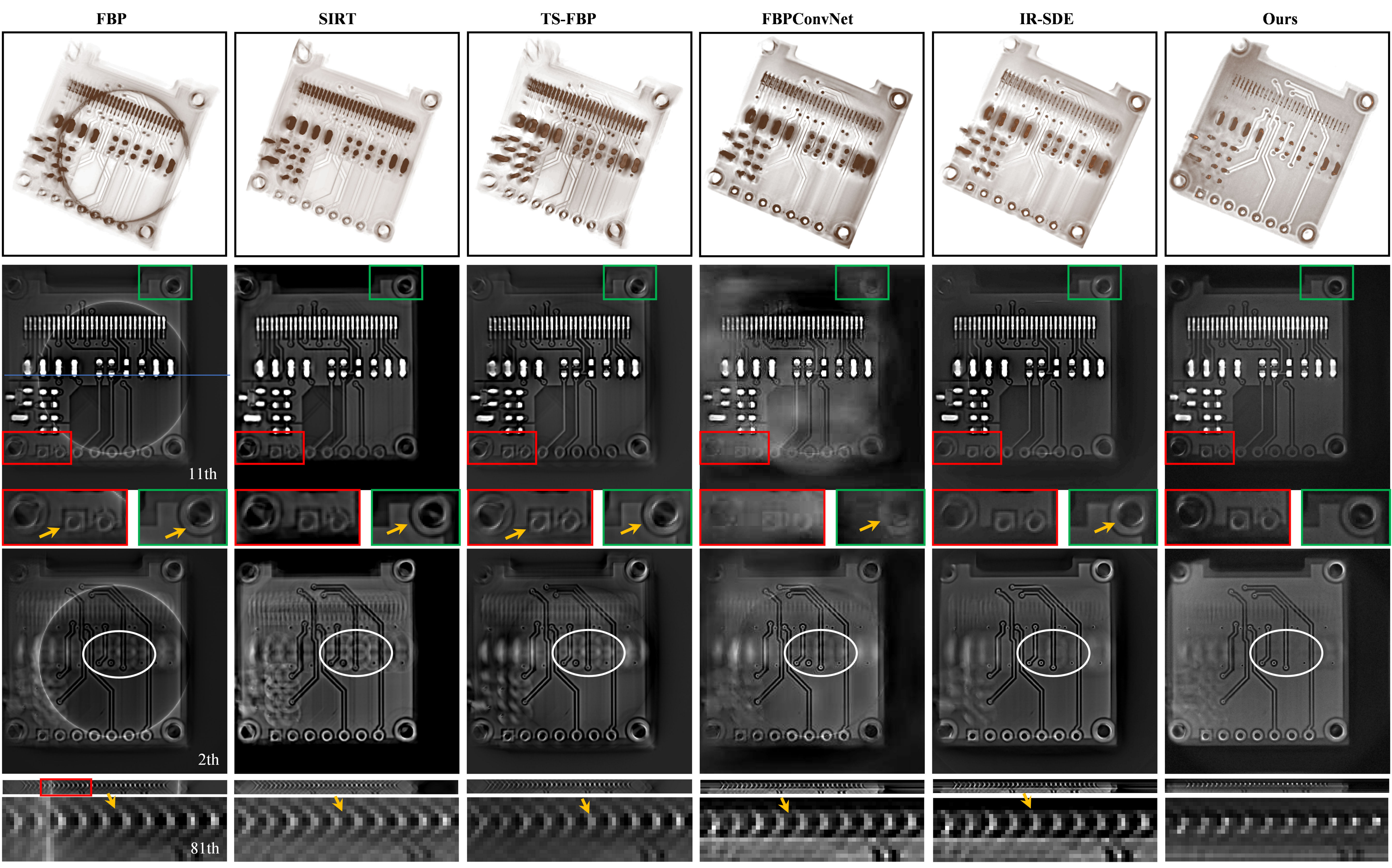}}
\caption{Qualitative results of real-world non-global truncated data. From top to bottom: 3D rendering results, two in-plane results, enlarged image (green and red boxes), and layered results. The window for the in-plane and layered results are [0,1]. From left to right: FBP algorithm reconstruction results, and various comparative algorithms; the last column shows the proposed algorithm.}
\label{rng}
\end{figure*}

Figure \ref{rng} shows representative results of removing non-global truncated artifacts using different methods for visual comparison. The SIRT results show obvious aliasing artifacts, as indicated by the white circles. The TS-FBP method results in significant blurring of the truncated outer region. The FBPConvNet method exhibits significant distortion with numerous fog-like artifacts. The IR-SDE method performs better than previous methods, indicating that the diffusion model has good generalization ability. However, structural distortion problems still exist, as shown by the yellow arrows in the ROI maps. Our method achieves the best structure recovery, demonstrating its good generalization ability, which is attributed to our model incorporating the truncated physical model. Furthermore, the yellow arrows in the layered diagrams show that all methods exhibit significant aliasing artifacts; in comparison, our method shows less aliasing artifacts due to our designed Cycle Spinning TI mechanism.

\subsubsection{Evaluation on Global Truncated Datasets} 
\begin{figure*}[pos=t]
\centerline{\includegraphics[width=0.9\linewidth]{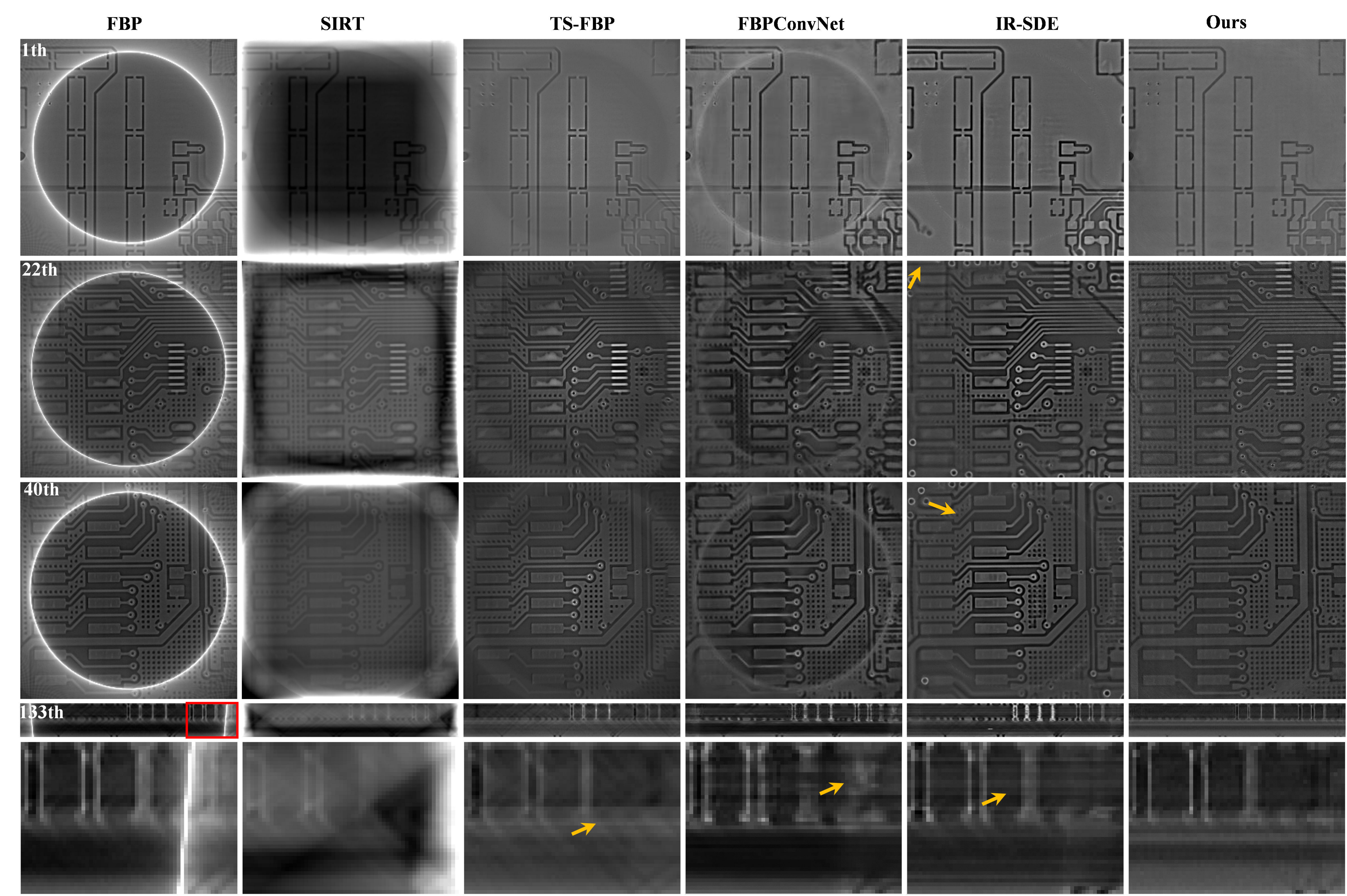}}
\caption{Qualitative results of real-world global truncated data. From top to bottom: Results from three horizontal planes and layered results. The window for the in-plane and layered results are [0,1]. From left to right: FBP algorithm reconstruction results, and various comparative algorithms; the last column shows the proposed algorithm.}
\label{rg}
\end{figure*}

Figure \ref{rg} shows representative results of removing global truncated artifacts using different methods for visual comparison. Note that, on the one hand, since we directly used the simulation model for the actual experiment, if we enlarge the reconstruction area of the actual experiment to be the same as the simulation experiment, i.e., to see all data incomplete areas, the image detail resolution will be extremely low when the reconstruction size is fixed at 256. On the other hand, the actual scanned object is much larger than the simulation, which also means we cannot expand the reconstruction range too much. Therefore, in this actual experiment, our reconstructed image only contains a portion of the data incomplete areas. The TS-FBP method still exhibits truncation artifacts, indicating its poor performance under severe truncation conditions. The FBPConvNet method also shows significant distortion, with residual truncation artifacts remaining. The IR-SDE method performs significantly better, but closer inspection reveals that the porous structures in the areas indicated by the yellow arrows are smoothed out or blurred. In comparison, our method performs best. Furthermore, the yellow arrows in the layering diagram show obvious interlayer discontinuities in FBPConvNet and IR-SDE, while our method exhibits better interlayer continuity. This is attributed to the incorporation of wavelet-based regularization in the z-direction.

\subsection{Ablation Studies}
\begin{figure}[pos=t]
\centerline{\includegraphics[width=\linewidth]{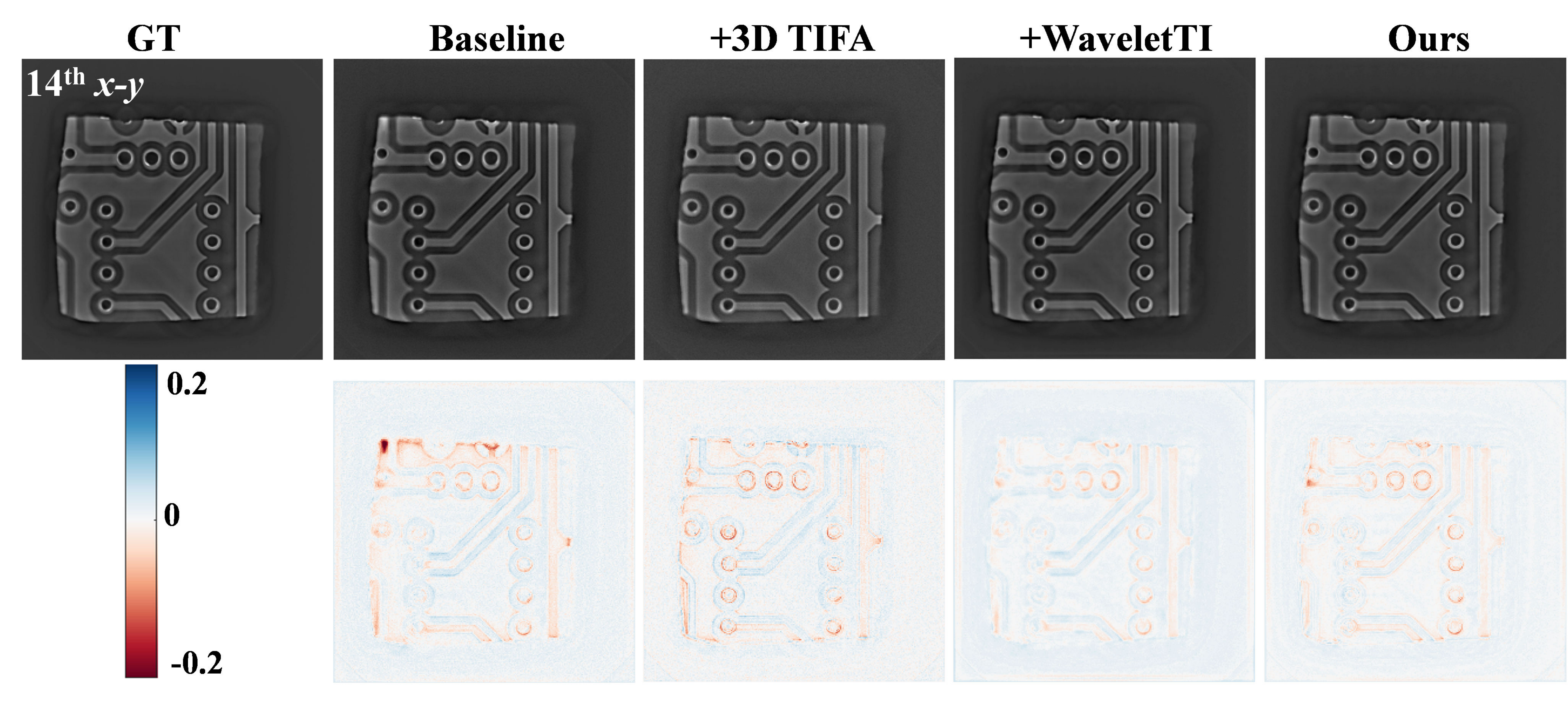}}
\caption{Visual evaluation results of the ablation study. The second row shows the difference maps with a window of [-0.2, 0.2].}
\label{ab2}
\end{figure}

We conducted ablation experiments to examine the impact of all components on the model results. All ablation experiments were performed on simulations with non-global truncation. For fair comparison, we used the original DiffusionMBIR \cite{chung2023solving} as the baseline. As show in Table. \ref{tab:ablation_study} the ablation study quantifies the contribution of each proposed component. While the 3D TIFA module provides a significant computational speedup (from 83 min to ~5 min), it introduces a noticeable drop in reconstruction fidelity. However, the integration of the Wavelet module effectively compensates for this loss, recovering 4.15 dB in PSNR with negligible time overhead. Our full configuration achieves a superior trade-off between efficiency and accuracy, outperforming the baseline in both metrics. The visual evaluation results shown in Fig. \ref{ab2} further demonstrate the effectiveness of the proposed module.

\begin{table*}[h]
\centering
\caption{Ablation Study Results}
\label{tab:ablation_study}
\begin{tabular}{cccccc}
\toprule
Baseline & 3D TIFA & Wavelet-TI & PSNR & SSIM & Time(s) \\
\midrule
$\surd$ &         &         & 38.40 & 0.9546 & 4980 \\
$\surd$ & $\surd$ &         & 35.63 & 0.9257 & 313 \\
$\surd$ &         & $\surd$ & \textbf{40.36} & \textbf{0.9800} & 5640 \\
$\surd$ & $\surd$ & $\surd$ & 39.78 & 0.9783 & 342 \\
\bottomrule
\end{tabular}
\end{table*}

\section{Discussion}
\subsection{Discussion on Regularization Direction}
\begin{figure}[pos=t]
\centerline{\includegraphics[width=\linewidth]{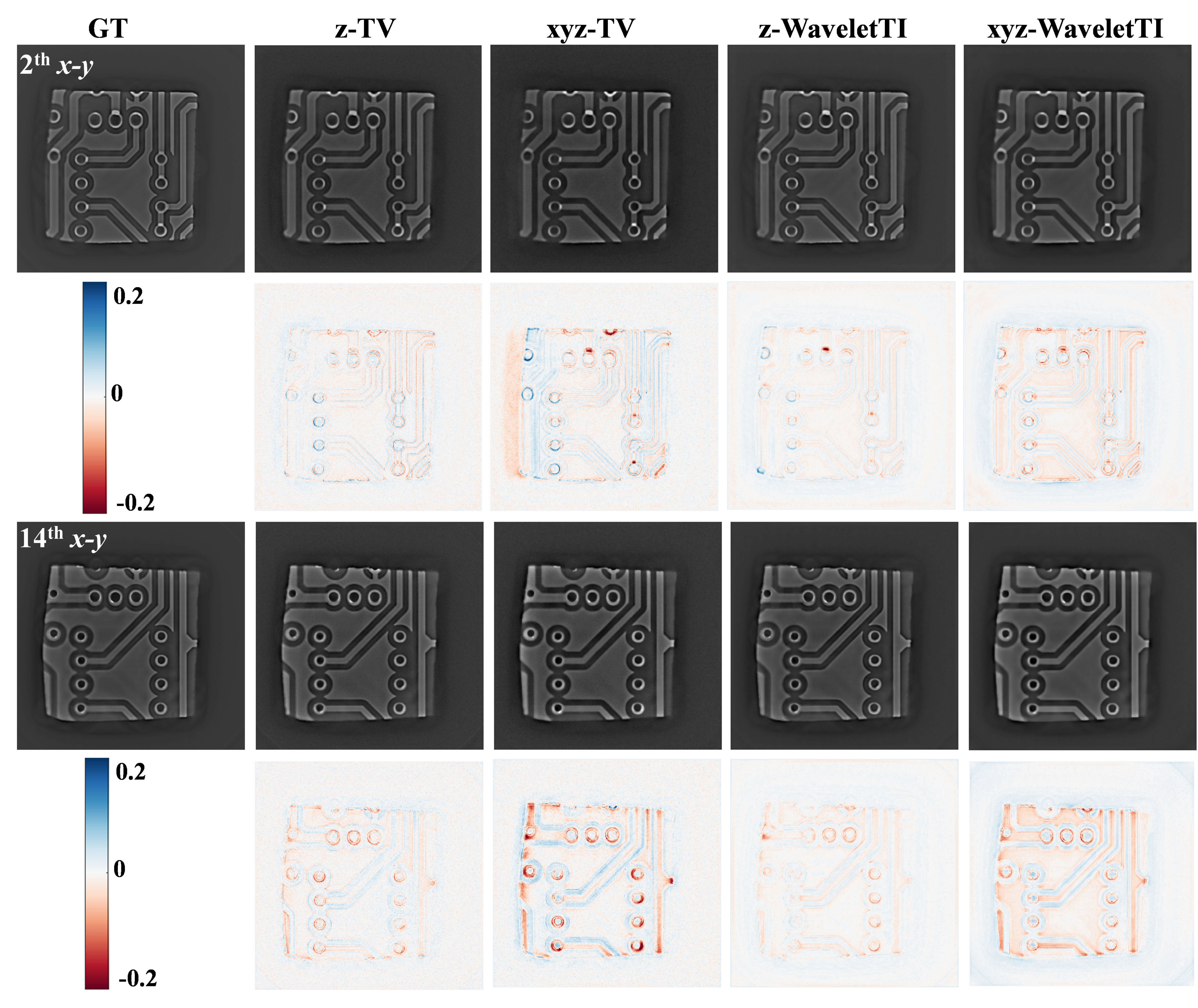}}
\caption{Visual evaluation results of regularization with different directions.
The second and fourth rows are difference maps with a window of [-0.2, 0.2].
z- and xyz- denote regularization in the z-direction and xyz three directions, respectively.}
\label{ab1}
\end{figure}

To verify the influence of multi-directional regularization versus uni-directional regularization on the results, we implemented z-direction TV regularization, xyz three-direction TV regularization, z-direction WaveletTI regularization, and xyz three-direction WaveletTI regularization separately.
The visual results are shown in the Fig. \ref{ab1}. From the difference maps, the three-directional regularization performs worse than the z-directional one, and the differences are mainly concentrated at the edges, indicating large discrepancies between the edges and the ground truth.
In addition, the Table \ref{tab:regularization_transposed} presents the quantitative evaluations of several regularization methods, which further demonstrate that the z-directional regularization is significantly superior to multi-directional regularization.

\begin{table}[htbp]
  \centering
  \caption{Comparison of 3D Image Evaluation Metrics for Different Regularization Methods}
  \label{tab:regularization_transposed}
  \begin{tabular}{lccc}
    \toprule
    Regularization Type & PSNR & SSIM & GMSD \\
    \midrule
    z-TV          & 35.63 & 0.9257 & 0.0247 \\
    xyz-TV        & 32.11 & 0.9059 & 0.0399 \\
    z-WaveletTI   & 39.78 & 0.9783 & 0.0178 \\
    xyz-WaveletTI & 35.14 & 0.9678 & 0.0272 \\
    \bottomrule
  \end{tabular}
\end{table}



\subsection{Discussion on Advantages and Disadvantages of Our Method}
Experimental results show that the proposed method outperforms other competing models both quantitatively and qualitatively, and exhibits promising potential in artifact reduction and data extrapolation. We discuss our advantages from the following four aspects:
1) We integrate measurement data into the sampling of the diffusion model via a model-based iterative reconstruction algorithm, which enhances data consistency and interpretability compared with end-to-end learning.
2) We elegantly solve the 3D CL truncation problem using a method that combines 2D priors with z-directional regularization, achieving superior performance on a slice-by-slice basis compared with 2D methods.
3) By combining the translation-invariant principle with wavelet regularization, we obtain better performance than TV regularization.
4) We apply a time-backtracking sampling strategy to our model, which greatly improves sampling efficiency. Meanwhile, the combination of coarse and fine sampling realized by time backtracking ensures reconstruction quality, which is crucial for industrial scenarios requiring high efficiency.

However, we also acknowledge several limitations of the proposed method.
First, although we have significantly improved sampling efficiency compared with conventional diffusion models, the current sampling speed is still insufficient for practical industrial scenarios. In the future, we plan to further accelerate inference efficiency using strategies such as flow matching \cite{lipman2024flow} and consistency models \cite{song2023consistency}.
Second, aliasing artifacts, which are unique to CL scanning, still exist. Although the proposed method can effectively remove truncation artifacts, it lacks suppression of aliasing artifacts. Aliasing artifacts essentially stem from the limited-angle problem. In the future, we plan to integrate regularization strategies with aliasing suppression into the proposed model to simultaneously remove truncation artifacts and alleviate aliasing.

\section{Conclusion}
This paper proposes a novel 3D accelerated diffusion model for CL truncation reconstruction. A primary achievement of this work is its capability to recover data-incomplete regions, which effectively expands the imaging FOV and greatly enhances scanning efficiency. To support this extended reconstruction, the model employs four key techniques: (i) a pseudo-3D strategy combining a 2D diffusion model with slice aggregation to ensure strict 3D data consistency; (ii) $z$-directional wavelet regularization to mitigate inter-slice discontinuities; (iii) a cycle-spinning TI mechanism paired with a low-frequency preservation strategy to suppress aliasing artifacts; and (iv) a customized 3D fast-sampling architecture. Simulation and practical experimental results demonstrate the superiority and effectiveness of our method in eliminating severe truncation artifacts across the expanded FOV. Notably, our proposed method exhibits high efficiency and fidelity, making it a highly promising solution for the practical industrial inspection of PCBs.






\bibliographystyle{IEEEtran}
\bibliography{ref}

\section{Appendix}
\subsection{Detailed Derivation of Sampling Region Boundaries} 
\label{a}
To analyze the sampling characteristics in truncated scanning, a geometric equivalent transformation is employed, where the object remains stationary while the X-ray source and detector perform a relative circular motion around it[cite: 26]. The geometric center of the PCB is set as the origin $O(0,0,0)$, and any voxel point $(x,y,z)$ is represented in polar coordinates as $(r, \theta, z)$, where $r=\sqrt{x^2+y^2}$.

\subsubsection{Initial Geometric Parameters}
The opening angle $\delta$ of the X-ray beam relative to the central ray is determined by the detector width and the source-to-detector distance $S_{DD}$:
\begin{equation}
    \delta = \arctan\left(\frac{m \cdot d}{2 \cdot S_{DD}}\right)
\end{equation}
where $m$ is the number of detector elements and $d$ is the pixel width[cite: 17, 20]. The z-coordinate of the X-ray source is $z_{source} = -S_{OD} \cdot \sin\varphi$[cite: 22]. When the ray reaches a slice at height $z$, the total vertical drop $\Delta Z$ is given by:
\begin{equation}
    \Delta Z = z - z_{source} = z + S_{OD} \cdot \sin\varphi
\end{equation}

\subsubsection{Derivation of Boundary Radii}
The complete sampling region (CSR) at height $z$ is constrained by three physical boundaries:

\begin{enumerate}
    \item \textbf{Upper half-cone constraint $R_{up}(z)$}: This boundary is defined by the innermost ray of the light cone on the same side. With an inclination angle of $(90^{\circ}-\varphi-\delta)$, the radius is calculated by subtracting the radial contraction due to the ray's ascent from the source's rotation radius:
    \begin{equation}
        R_{up}(z) = S_{OD} \cdot \cos\varphi - (z + S_{OD} \cdot \sin\varphi) \cdot \tan(90^{\circ} - \varphi - \delta)
    \end{equation}

    \item \textbf{Lower half-cone constraint $R_{dn}(z)$}: This is limited by the outermost ray from the contralateral scanning position (180°), with an inclination angle of $(90^{\circ}-\varphi+\delta)$:
    \begin{equation}
        R_{dn}(z) = (z + S_{OD} \cdot \sin\varphi) \cdot \tan(90^{\circ} - \varphi + \delta) - S_{OD} \cdot \cos\varphi
    \end{equation}

    \item \textbf{Transverse field of view constraint $R_{trans}(z)$}: For a point $(r,\theta,z)$, its transverse projection coordinate $u(\theta)$ on the tilted detector is:
    \begin{equation}
        u(\theta) = S_{DD} \cdot \frac{r \cos\theta}{(S_{OD} + z \cdot \sin\varphi) + r \sin\theta \cos\varphi}
    \end{equation}
    To find the maximum projection distance, we set the derivative $\frac{du(\theta)}{d\theta} = 0$, which yields the condition $\sin\theta = -\frac{r \cos\varphi}{S_{OD} + z \cdot \sin\varphi}$[cite: 38]. Substituting this back into the expression for $u(\theta)$, the analytical solution for the maximum projection distance $|u|_{max}$ is:
    \begin{equation}
        |u|_{max} = \frac{S_{DD} \cdot r}{\sqrt{(S_{OD} + z \cdot \sin\varphi)^{2} - (r \cos\varphi)^{2}}}
    \end{equation}
    By setting $|u|_{max} = \frac{m \cdot d}{2}$, the transverse truncation boundary $R_{trans}(z)$ is derived as:
    \begin{equation}
        R_{trans}(z) = \frac{\frac{m \cdot d}{2} \cdot (S_{OD} + z \cdot \sin\varphi)}{\sqrt{S_{DD}^2 + (\frac{m \cdot d}{2} \cdot \cos\varphi)^2}}
    \end{equation}
\end{enumerate}

\subsubsection{Incomplete Sampling Boundary}
The outer boundary of the incomplete sampling region is defined by the maximum absolute outer envelope surface $R_{sweep}(z)$ formed by the sweeping ray cone:
\begin{equation}
    R_{sweep}(z) = S_{OD} \cdot \cos\varphi + (z + S_{OD} \cdot \sin\varphi) \cdot \tan(90^{\circ} - \varphi + \delta)
\end{equation}

\subsection{ADMM-Wavelet Solution}
\label{b}

In this section, we will derive the three optimization subproblems of ADMM-Wavelet.

{\bf{The $\mathbf{x}$-Subproblem:}} In the $k$-th iteration of the Alternating Direction Method of Multipliers (ADMM), the variables $\mathbf{z}_{\text{wave}}^{(k)}$ and $\mathbf{u}_{\text{wave}}^{(k)}$ are fixed. The optimization subproblem with respect to $x$ is formulated as:
\begin{equation}
    \mathbf{x}^{(k+1)} = \arg\min_{\mathbf{x}} \frac{1}{2}\|\mathbf{y} - \mathbf{A}\mathbf{x}\|_{2}^{2} + \frac{\rho}{2}\|\Psi_{\mathbf{z}}(\mathbf{x}) - \mathbf{z}_{\text{wave}}^{(k)} + \mathbf{u}_{\text{wave}}^{(k)}\|_{2}^{2}
\end{equation}
This formulation is a standard unconstrained convex quadratic optimization problem. Taking the first-order partial derivative with respect to $\mathbf{x}$ and setting it to zero yields:
\begin{equation}
    \mathbf{A}^{T}(\mathbf{A}\mathbf{x} - \mathbf{y}) + \rho\Psi_{\mathbf{z}}^{T}(\Psi_{\mathbf{z}}\mathbf{x} - \mathbf{z}_{\text{wave}}^{(k)} + \mathbf{u}_{\text{wave}}^{(k)}) = 0
\end{equation}
Expanding the terms and rearranging the equation produces a linear system in standard form:
\begin{equation}
    (\mathbf{A}^{T}\mathbf{A} + \rho \mathbf{I})\mathbf{x}^{(k+1)} = \mathbf{A}^{T}\mathbf{y} + \rho\Psi_{\mathbf{z}}^{T}(\mathbf{z}_{\text{wave}}^{(k)} - \mathbf{u}_{\text{wave}}^{(k)})
\end{equation}
Due to the orthogonality of the Haar wavelet operator ($\Psi_{\mathbf{z}}^{T}\Psi_{\mathbf{z}} = \mathbf{I}$, the identity matrix), the initially complex covariance matrix significantly simplifies to:
\begin{align}
    \mathbf{A}_{CG} &= \mathbf{A}^{T}\mathbf{A} + \rho \mathbf{I} \\
    b_{CG}^{(k)} &= \mathbf{A}^{T}\mathbf{y} + \rho\Psi_{\mathbf{z}}^{T}(\mathbf{z}_{\text{wave}}^{(k)} - \mathbf{u}_{\text{wave}}^{(k)})
\end{align}

The resulting coefficient matrix $\mathbf{A}^{T}\mathbf{A} + \rho \mathbf{I}$ exhibits excellent diagonal dominance and symmetric positive definite properties. Consequently, we employ the Conjugate Gradient (CG) method to solve this linear system. In the implementation, the projection residual $\mathbf{r} = b_{CG} - \mathbf{A}_{CG}\mathbf{x}$ is computed, and the optimal descent direction $p$ is searched within the Krylov subspace. Because of the substantial improvement in the condition number of the Hessian matrix, the CG algorithm typically requires very few inner iterations (e.g., $N_{\text{inner}} = 2$) to achieve high computational precision. This provides a revolutionary efficiency advantage over the dozens of CG iterations traditionally required by Total Variation (TV) regularization.

{\bf{The $\mathbf{z}$- and $\mathbf{u}$-Subproblems:}} Following the update of $\mathbf{x}^{(k+1)}$, the variables $\mathbf{x}$ and $\mathbf{u}_{\text{wave}}$ are fixed to solve the $\mathbf{z}_{\text{wave}}^{(k)}$ subproblem:
\begin{equation}
    \mathbf{z}_{\text{wave}}^{(k+1)} = \arg\min_{\mathbf{z}} \lambda_{1}\|\mathbf{z}\|_{1} + \frac{\rho}{2}\|\Psi_{z}(\mathbf{x}^{(k+1)}) - \mathbf{z} + \mathbf{u}_{\text{wave}}^{(k)}\|_{2}^{2}
\end{equation}
According to proximal operator theory, quadratic optimization problems incorporating $L_{1}$ regularization possess a closed-form solution. This solution can be precisely obtained using the soft-thresholding operator $\mathcal{S}_{\kappa}$:
\begin{equation}
    \mathbf{z}_{\text{wave}}^{(k+1)} = \mathcal{S}_{\lambda_{1}/\rho} \left( \Psi_{\mathbf{z}}(\mathbf{x}^{(k+1)}) + \mathbf{u}_{\text{wave}}^{(k)} \right)
\end{equation}
The soft-thresholding operator is mathematically defined as $\mathcal{S}_{\kappa}(a) = \text{sign}(a) \cdot \max(|a| - \kappa, 0)$. Finally, the dual variable $\mathbf{u}_{\text{wave}}$ is updated based on gradient ascent:
\begin{equation}
    \mathbf{u}_{\text{wave}}^{(k+1)} = \mathbf{u}_{\text{wave}}^{(k)} + \Psi_{\mathbf{z}}(\mathbf{x}^{(k+1)}) - \mathbf{z}_{\text{wave}}^{(k+1)}
\end{equation}
This update of the dual variable penalizes the discrepancy between the physically consistent solution and the prior wavelet manifold, driving the entire ADMM algorithm to converge to a global saddle point.

\vfill

\end{document}